# Searching for chromate replacements using natural language processing and machine learning algorithms

S. Zhao[1,2*], N. Birbilis[1,2*]

[1]College of Engineering and Computer Science, The Australian National University, Acton, A.C.T., 2601, Australia.
[2]ARC Research Hub for Australian Steel Innovation

*shujing.zhao@anu.edu.au
*nick.birbilis@anu.edu.au

**Abstract**

The past few years has seen the application of machine learning utilised in the exploration of new materials. As in many fields of research – the vast majority of knowledge is published as text, which poses challenges in either a consolidated or statistical analysis across studies and reports. Such challenges include the inability to extract quantitative information, and in accessing the breadth of non-numerical information. To address this issue, the application of natural language processing (NLP) has been explored in several studies to date. In NLP, assignment of high-dimensional vectors, known as embeddings, to passages of text preserves the syntactic and semantic relationship between words. Embeddings rely on machine learning algorithms and in the present work, we have employed the Word2Vec model, previously explored by others, and the BERT model – applying them towards a unique challenge in materials engineering. That challenge is the search for chromate replacements in the field of corrosion protection. From a database of over 80 million records, a down-selection of 5990 papers focused on the topic of corrosion protection were examined using NLP. This study demonstrates it is possible to extract knowledge from the automated interpretation of the scientific literature and achieve expert human level insights.

## Nomenclature

*Abbreviations*

| | |
|---|---|
| BERT | Bidirectional Encoder Representations from Transformers |
| CLS | A 'classification token'. A sentence-level token introduced by BERT. |
| MASK | A MASK is a token, used to provide a token where a model can predict by "filling the blanks" |
| MLM | Masked Language Modelling |
| NLP | Natural Language Processing |
| SEP | A 'special token' in BERT |

## Introduction

The corrosion of metals and alloys remains a significant technological and financial issue - globally. Studies regarding the cost of corrosion, for example, those recently conducted in the USA[1], China[2], and Australia[3], reveal that corrosion costs amount to ~3% of GDP annually (which equates to global costs of > US$ 1 trillion per annum). As a result, methods for corrosion prevention remain critical. In terms of the corrosion protection of metals and alloys, for over half a century, the benchmark for exceptional performance from corrosion inhibiting compounds has been demonstrated by hexavalent chromium (known as chromate) [4]. Chromate is a powerful inhibitor, as it can passivate many metals, including Zn, Al, Mg, etc. The mechanism of chromate protection involves the formation of a protective Cr (III) oxide layer on reactive metals from mobile and soluble Cr (VI) oxyanions that can migrate to 'active' (anode) sites [5]. Chromate serves as a corrosion inhibitor in aqueous solutions, but also as an additive to primers used to coat metals (such as steels and galvanised steels).

The International Agency for Research on Cancer (IARC) confirmed that hexavalent chromium (Cr (VI)) is a human carcinogen in 1990 based on independent studies around the world [6]. However, the corrosion inhibition performance of chromate containing primers is appreciable, such that chromate-containing primers are the current industry benchmark in terms of performance (and additionally, consumer expectations of product performance). Given the documented concerns regarding the use of chromate and its disposal [7-9], the evolution towards chromate free corrosion inhibitors is underway. In a tangible sense, there are already numerous chromate free corrosion inhibition strategies utilised in consumer products today. The adoption to alternative (chromate free) approaches is progressing as suitable alternatives to chromate are identified albeit few are as: (i) cost effective, (ii) passivating, and (iii) applicable across a wide range of metals and alloys, as chromate.

A review by Gharbi and co-workers into chromate alternatives, summarised that singular alternatives to chromate as a 'drop in' replacement strategy are unlikely. The past three decades have seen much research focus on alternatives for chromates. Some of the most widely explored alternatives that have demonstrated promising performance approaching that of chromate-containing inhibitors include rare-earth-based inhibitors [10] and rare-earth coatings, vanadate-based coatings [11] that are currently utilised in aerospace systems [12, 13], lithium-containing coatings [14] organic coatings, nanocomposites, phosphate coatings [15] and metal rich primers [16]. Undoubtedly, the search for chromate alternates remains a very timely topic, and a puzzle that is yet to be solved in the field.

The rapidly growing and large-scale material science knowledge base is typically published as archival 'papers'. In this content, text mining has been one of the most exciting tools in recent years [17-20]. Most literature text remains unstructured or semi-structured data (natural language) which is not capable of being readily interpreted by computer (whereby a computer is unable to readily interpret context). However, to extract comprehensible and meaningful information from text, supervised natural language processing (NLP) and machine learning methods have been shown to be promising, and resulted in the exploration of text mining in the field of material science [21-24]. Supervised NLP requires part of the corpus (i.e., a body of writing) to be in the form of human-annotated data for training, and then tested by unlabelled text. Some supervised NLP algorithms include Support Vector Machines (SVM), Bayesian Network (BN), Maximum Entropy (ME), Conditional Random Field (CRF), as well as several other algorithms [25-28]. However, the vast majority - if not essentially all the open literature reports and published data - are unlabelled. Therefore, such text and data may be mined by unsupervised NLP algorithms. Clustering, a well-known unsupervised machine learning algorithm of classifying similar data into groups, has been demonstrated and used to generate machine learning datasets, and to identify noisy data, in material science [29, 30].

Tshitoyan, Dagdelen [31] utilised Word2Vec, an unsupervised word embedding method, to extract underlying structure-property relationships in materials and predict new thermoelectric materials [18]. Word2Vec is a vector representation of words, which allows similar words to have a similar representation. Although word embedding is one of the most widely used representations of vocabulary, such an approach can only generate one vector for each word. Therefore, Word2Vec models are context-independent and different contexts of one word are not able to be taken into account. In addition, the Word2Vec model is not capable of learning Out-of-Vocabulary (OOV), since it generates tokens on the "word" level. However, given the promise that Word2Vec has shown to date, in the field of materials, in this study we apply the method towards the open literature in order to seek possible chromate alternatives. In addition to using Word2Vec, we have also explored the utilisation of a state-of-the-art model, known as Bidirectional Encoder Representations from Transformers (BERT).

BERT is a language representation model developed by Google in 2018, enabling pre-training deep bidirectional representations from unlabelled text, by jointly conditioning both the left and right context (outlined further below) in all Transformer encoder layers [32]. In the BERT model, sub-word tokenization is utilised, with a principle that rare words are decomposed into sub-words; whilst frequent words should not be split [32]. This allows the model to process words it has never seen before; meaning BERT is capable of learning Out-of-Vocabulary. The BERT tokenizer is based on WordPiece embedding with 30,000 tokens [32, 33], by implementing the following methods: i) Tokenizing (splitting texts to sub-word tokens), switch tokens to integers, and encoding/decoding; ii). Generating new tokens to the corpus; and iii). Adding and assigning special tokens (MASK, SEP and CLS).

A commonly used procedure for training models for various tasks in modern NLP systems, is to first pre-train a general model on large amount of unlabelled data, then finetune on downstream NLP tasks including classification, summarization, etc. Masked Language Modelling (MLM) is a pre-training method, and utilised for how BERT is pre-trained. Taking a sentence, the model randomly masks 15% of the words in the input, then runs the entire masked sentence through the model to predict the masked words [34]. This is different from traditional recurrent neural networks (RNNs) that usually see the words one after the other, or from autoregressive models like Generative Pre-trained Transformer (GPT) that every tokens only attends to context to its left, which internally masks future tokens [35]. The MLM approach allows the model to learn a bidirectional representation of a sentence. Figure 1 shows how MLM works, the input sentence is first tokenized with special tokens and fed into BERT as a sequence. The pre-training data is 800M words from BooksCorpus and 2,500M words from Wikipedia [32].

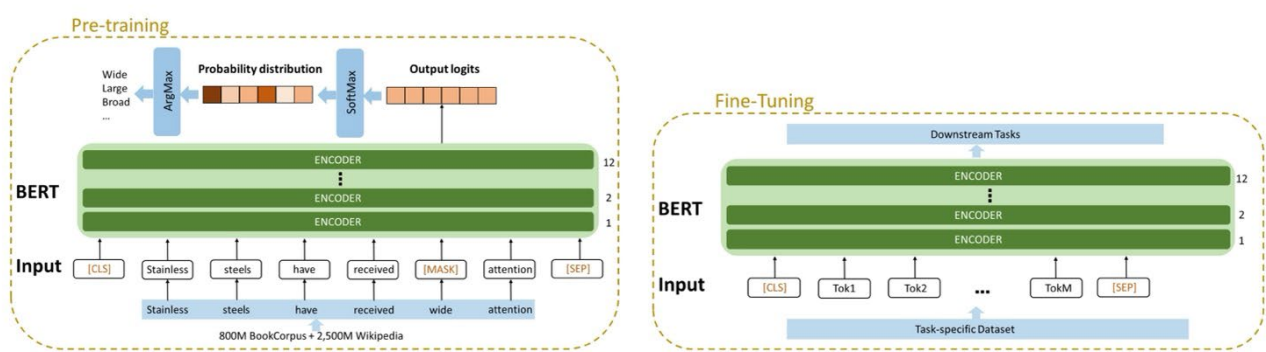

*Figure 1.* Schematic representation of Masked Language Modelling (MLM) with BERT, including pre-training and fine-tuning steps. The number of Transformer encoder layers is L, the hidden size is H, the number of self-attention heads as A. $BERT_{base}$ has a model size of: L=12, H=768, A=12, Total Parameter=110M.

The representation of every other input word can be weighted by α (attention weight) during learning MASK word. For example, α = 1 means that each other word has equal weight in the representation. The tokens are passed to Transformer encoder layers, each layer applies bidirectional self-attention. Inputs then pass through a feed-forward network, then to the next encoder layer. Each output logit is the size of the vocabulary size, and is transferred to a probability distribution by applying the softmax function [32]. A softmax function (Eqn. 1) is a normalization process that transforms K input values into K values between 0 to 1 that sums to 1.

$$\sigma(\vec{z})_i = \frac{e^{z_i}}{\sum_{j=1}^{K} e^{z_j}} \quad (1)$$

The output values then can be interpreted as probabilities. Predicted tokens are then calculated by applying argmax to probability distribution. An argmax function (Eqn. 2) is a function that returns the argument where the function has maximum value. Given a function, $f: X \rightarrow Y$, the argmax over subset $S$ of $X$ is defined as:

$$argmax_S\, f(x) := \{x \in S: f(s) \leq f(x)\ for\ all\ s\epsilon\ S\} \quad (2)$$

The argmax is used to locate the token/class with the largest predicted probability. In the case shown in Figure 1, for example, the predicted results could be "wide, large, broad…" [36]. The fine-tuning part has similar architecture with pre-training: For different downstream tasks, feed the model with task-specific inputs, add a task- specific output layer and finetune parameters.

## Methods

### Data collection and pre-processing

A total of 5990 entries were collected by accessing and extracting 84 million records from Scopus application programming interfaces (APIs) (https://dev.elsevier.com/). A set of wild card query terms was introduced to limit acquisition primarily relate to the relevant topic. Only articles with "chrom*"and "replace*" or "substitute" in their titles, abstracts or keywords were collected. Furthermore, abstracts were filtered through applying query terms "alumin*", "zinc", "magnesium", "alloy", "steel" or "iron" (to ensure that they were relevant to substrates of interest). Abstracts that were in non-English languages were removed from the corpus to allow the use of a singular language setting as English. A number of articles with copyright limitations or missing passages were also removed, as were articles with content types not corresponding to peer reviewed publications – leaving 1812 works forming the training dataset for the Word2Vec and BERT architecture.

Preprocessing of body text involved removing XML format and XML quotes tags, leading words such as "Abstract" were also eliminated. In the Word2vec model, we followed the general preprocessing steps as per the unsupervised word embedding study from 2019 [31]. Element and element names, numbers and units were converted to tokens such as #element, #nUm, #unit respectively. Material formulas were normalized in an alphabetical way, such that any chemical formula was simplified regardless of the order of elements. The processed dataset includes one study in each line, and it was tokenized to a combination of individual words through ChemDataExtractor [37]. Chemical formulas were recognized by applying pymatgen [38], regular expression and rule-based techniques, jointly. The body text was transformed to lowercase if the token was not a chemical formula or an abbreviation. Abbreviations were identified by instances when not only the first letter was uppercase. In the pre-trained BERT-based model, subword tokenization was used allowing the model to process words it has never seen before. The tokenizer includes 250,000+ tokens from the chemical domain. Rare words were tokenized into meaningful subwords while frequent words were only split into word tokens. BERT takes the whole input as a single sequence. Special tokens [CLS] and [SEP] were used to understand the input sequence. Besides token embeddings, BERT includes more information for each token with positional embeddings and segment embeddings.

### Training

In the Word2vec training, the same gensim model was utilised, following the hyperparameter tuning process performed on 14042 material science analogy pairs, as shown in Tshitoyan's work [31]. Hyperparameters are substantial parameters that control the learning process and evaluated prior to the training. Common hyperparameters include learning rate, optimization method, loss function, number of hidden layers, batch size, and epochs. Batch size is the length of data samples for training before the gradient descent updates. A development dataset for hyperparameter tuning process was created herein, in which 10% of data was extracted randomly from the original dataset. The optimization process is a grid search, and searches through a specified set of parameters in the hyperparameter space. Models were trained with each pair hyperparameters and evaluated by analogy score, defined as the rate of correctly matched analogies from two chemical and element name pairs. A set of optimal hyperparameters were gained with highest analogy score: a learning rate of 0.001, a size of embedding of 300, a batch size of 128 and 30 epochs. The training then performed by applying the set of optimal hyperparameters.

To focus the study in the Chemistry Domain, we used the pretrained chemical-bert-uncased model in Hugging Face (https://huggingface.co/recobo/chemical-bert-uncased), and finetuned the training based on the mask language modeling implementation in Huggin Face with some modifications [39]. This pretraining and finetuning process is shown in Figure 2. The chem-bert-uncased model is pretrained from SciBERT (https://huggingface.co/allenai/scibert_scivocab_uncased) with over 40,000 technical documents from Chemical Industrial and over 13,000 Wikipedia Chemistry articles. The software 'Wandb' (for tracking wights and biases) was introduced to track and visualize the hyperparameter tuning process [40]. Similarly, the best hyperparameters for training were selected by training a model on a small parcel of data (the development set) over each pair hyperparameter and calculating corresponding perplexity. Perplexity (given by Eqn. 3) is a commonly used value to evaluate language models in NLP:

$$PP(W) = 2^{H(W)} = 2^{-\frac{1}{N}\log_2 P(w_1, w_2, \ldots, w_N)} \qquad (3)$$

where H is the cross-entropy, P is the language model, w is a sequence of words, and N is length of the words

A lower perplexity commonly indicates a better language model. The hyperparameters pairs are epoch = (10,20,30), batch size = (16,32), learning rate = ($1e^{-5}$, $1e^{-4}$, $1e^{-3}$). As shown in Figure 3, the best perplexity corresponds to the optimal hyperparameter pair (epoch = 10, batch size = 32, learning rate = $1e^{-4}$). We then finetuned the model with this optimal hyperparameter pair on the processed abstract dataset.

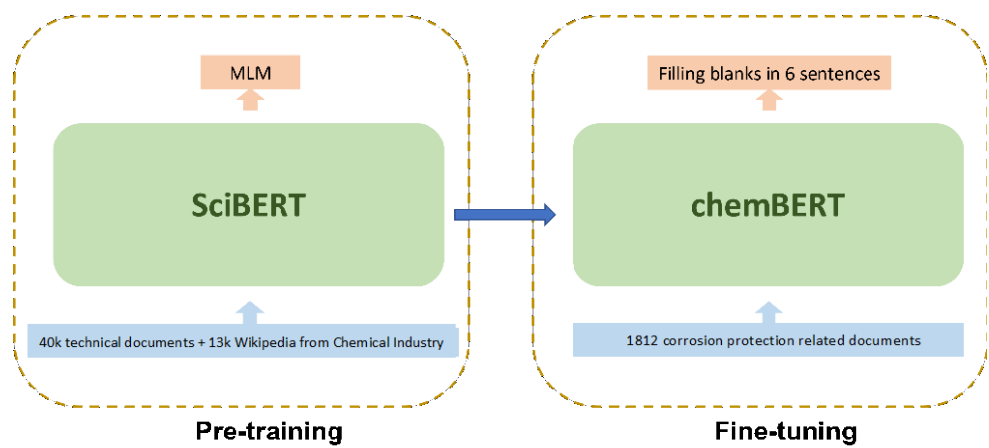

Figure 2. The pre-training and fine-tuning of BERT model in this study. The chemBERT model was pre-trained on the SciBERT model (designed for scientific texts) with technical documents and Wikipedia from the Chemical Industry. We then finetuned the chemBERT model with corrosion related documents.

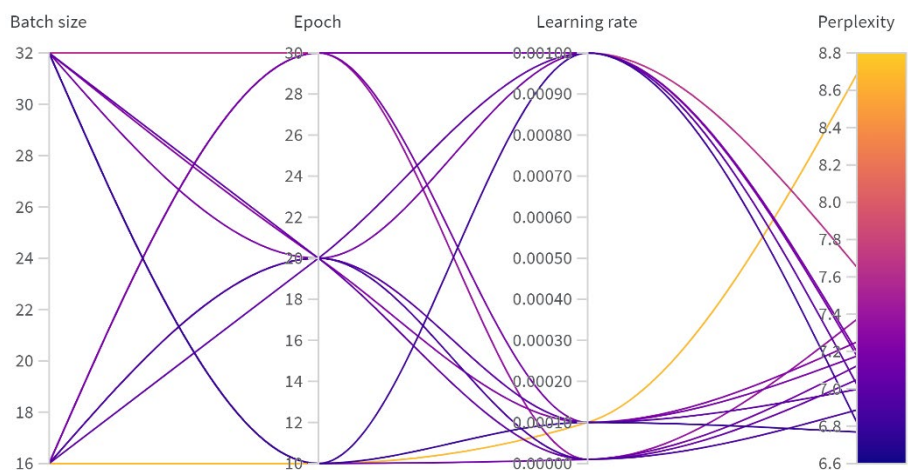

Figure 3. Hyperparameter tuning in BERT model to find the best hyperparameters for training. Each hyperparameter set (batch size, epoch and learning rate) was trained on a development dataset (10% of the whole dataset). The lower perplexity shows a darker purple color, indicates better performance of the language model with certain pair of hyperparameters.

### Evaluation

NLP tasks generally can be validated with measures such as accuracy, f-score, Root Mean Squared Error (RMSE), etc. However, the evaluation of unsupervised learning can be challenging due to the unlabelled output. This is primarily because common evaluation methods require comparing an output value against a known value. In the Word2Vec study, the cosine distance to vector "chromate" was used, to represent the probability of a material being a chromate replacement. That is, the chromate replacements are among the materials most similar to chromate, which were determined by the projection of normalized word embeddings. While in the

BERT experiment, we attempted to identify chromate replacements by "filling blanks" in a sentence. For example, top predictions of potential chromate replacements were sought by filling a [mask], whereby an example is: "Chromate can be replaced by [mask]". Our model randomly masks 15% of the input, runs the whole masked sentence, and outputs prediction of the masked words. Both the predicted results were categorized and compared with a benchmark alternative list which is summarized from known alternative corrosion preventative technologies, as discussed in the Results. The six masked sentences fed into the BERT model to seek potential alternatives to chromate are listed in Table 1.

Herein, the alternatives to chromate predicted were refined to a list of twenty categories of benchmark alternatives, as described in the Results section.

Table 1. The six masked sentences designed to seek potential alternatives for chromate in the BERT model.

| # | Masked sentence | Abbreviation |
|---|---|---|
| 1 | hexavalent chromium can be replaced by [MASK] | can |
| 2 | hexavalent chromium may be replaced by [MASK] | may |
| 3 | chromate can be replaced by [MASK] | chromate |
| 4 | the best corrosion inhibitor is [MASK] | inhibitor |
| 5 | [MASK] performed better than chromate | perform |
| 6 | the best conversion coating is [MASK] | coating |

## Results

Based on the results from previous work [31] an advantageous outcome from the Word2Vec representation was identified as the ability to represent both 'application words' and 'material formulae' similarly. In the present work, we therefore expect that the cosine distance of materials that have the same application, to be close; i.e. when the cosine similarity of a material representation and "chromate" representation is high, such materials are very likely to have similar applications to chromium and therefore candidate alternatives to chromate. Therefore, the closeness of the cosine distance is utilised for the determination of candidate alternatives to chromate when using the Word2Vec model.

In the utilization of the BERT MLM model, the different mode of operation of the model was exploited in order to seek chromate alternatives by posing open questions – namely six questions that required filling by six different masks. The top predictions arising from the exploration of the six [MASK], were deemed the candidate alternatives for chromate replacements.

For Word2Vec model and each of the six masked BERT model, the top 1000 results generated from the models were extracted, and this list of the top 1000 results was sorted to identify the candidate alternatives that are actually materials (i.e. materials, chemicals, or compounds) and that are relevant to the corrosion domain. Whilst most of the top 1000 results were related to the topic of materials, corrosion and alloys (to a large extent), we excluded common terms that had no relevance as alternates (i.e. 'steel') and any terms that were not materials (i.e. general conversational words). The number of relevant suggestions for chromate alternatives from the Word2Vec and BERT models, that are highly related to corrosion protection, are shown in Figure 4.

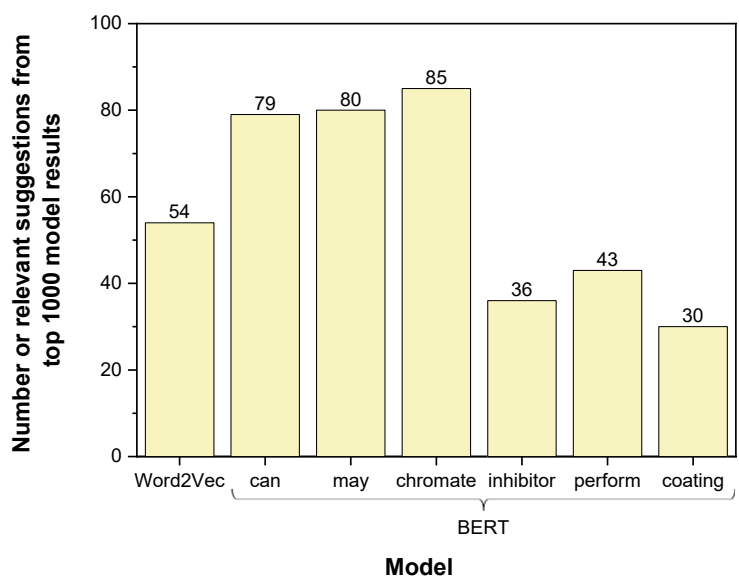

*Figure 4.* Number of relevant suggestions that were corrosion protection strategies from the top 1000 results generated by the respective Word2Vec and BERT models.

Of the top 1000 entries generated from each model, the Word2Vec approach detected 54 materials (which is inclusive of materials, compounds, or chemicals) that are relevant to serving as suitable alternatives to chromate. Conversely, the BERT approach identified a number of suitable alternatives that varied from 30 to 85 – depending on the question asked. From Figure 4, it is evident that the first three [mask] containing questions from the BERT model, yielded a relatively higher number of relevant results, which was correlated (by the authors) to how the mask containing sentence was structured. The BERT model sentences with the lowest yield include "The best corrosion inhibitor is [mask]" and "The best conversion coating is [mask]". These two sentences have the prospect of generating a large number of verbs as outputs, such as "obtained", or "needed", or, adjectives as outputs such as "available", or "possible" – all of which are reasonable to fill the sentence (but are not relevant materials). If we combine the results from six masked BERT models, the BERT model was capable of predicting 161 individual relevant suggestions for chromate alternatives. A list of each of the ranked chromate alternatives tallied in Figure 4, is presented in Appendix 1 and 2.

Of the chromate alternative results predicted by the Word2Vec model, some results were the same as the results from the BERT model, with an overlap of 19%. When interpreting the results obtained, it was observed that the results from Word2Vec have all appeared at least once in the corpus. However, the BERT model was not only able to identify some low frequency results, but also identified results that had never appeared in the corpus. For example, frequency of "cvd" (which is chemical vapour deposition) is one and its prediction rank is almost the same as "nanoparticles" which appears 61 times, as well as, "formaldehyde", "acrylate" etc. that do not exist in the dataset. This is important insight, because since the BERT model uses sub-word tokenization during its training, the model can "mix and match" (between the pre-training and fine-tuning), allowing the model to predict words not seen before in the corpus of corrosion protection relevant training data. The pre-training step, which was carried out using technical documents (from the SciBERT database) and Wikipedia (from the chemBERT database) – is where the BERT model would have seen such words – and is then able to use such words following fine-tuning. This indicates that the ability to pre-train BERT models using a vast array of less-specialist text is meaningful, as the BERT model is able to predict in a human-like manner (including out-of-field).

To illustrate how many alternatives have the potential for replacement of chromate, we compared the results of the predictions from the Word2Vec and BERT MLM model, with a list of benchmark chromate replacements. The list of potential alternatives was derived from three sources, each of which is a culmination of 'expert' level human analysis – and years of research and literature analysis [41-43]. The three studies/reports from which the benchmark list of chromate replacements was derived were not utilized in the model training herein, and reserved as an independent validation. The benchmark alternative list was curated into 20 categories, ranging from the trivalent chromium, rare earth-based coatings, vanadate-based coatings, Li-containing coatings, organic systems, phosphate-based systems to Mg-rich primers – as shown in Table 2.

When reviewing the outputs of the Word2Ve and BERT model, the authors manually identified relevant materials (suitable for consideration as chromate replacements) and allocated them to the benchmark category to which they relate – as also seen in Table 2.

Table 2. Benchmark alternatives to chromate with given prediction results examples.

| # | Benchmark alternatives to chromate [41-43] | Examples from the NLP predictions |
|---|---|---|
| 1 | Trivalent chromium | |
| 2 | Rare earth based inhibitors (aka lanthanide systems) | cerium, $CeN_3O_9$ |
| 3 | Vanadate based inhibitors | $BiO_4V$ ($BiVO_4$), vanadate |
| 4 | Li-containing conversion coatings / primers | lithium, LDH |
| 5 | Organic systems | polyurethane, amines, BTA |
| 6 | Phosphate-based systems | zinc phosphate, phosphates |
| 7 | Mg-rich primers | |
| 8 | Zirconium conversion coatings | |
| 9 | Titanium containing conversion coatings | |
| 10 | Silicon based systems | silanes and sol-gel |
| 11 | Zinc-based coatings | |
| 12 | Molybdate compounds | molybdate, $MoNa_2O_4$ ($Na_2MoO_4$) |
| 13 | Calcium based systems | |
| 14 | Electro-coatings / Electrophoretic systems | Electroplating, Co-P |
| 15 | Nanocomposites incl. nanoparticles | alumina, ceria, titania, graphene |
| 16 | TFSAA (thin film sulfuric acid anodisation) | anodising |
| 17 | BSAA (boric sulfuric acid anodising) | anodising, sulphuric |

| 18 | Filled electroless nickel | |
| 19 | WC/C (tungsten carbide carbon) coating | WC–Co, WC, (WCr)$_2$C-Ni |
| 20 | Nitride coatings | TiN, Cr$_x$N, TiN/CrN |

To analyze the efficiency of the NLP models to predict chromate alternatives in an automated manner, we summarized the number of benchmark related results in each category, for each MLP model and present the results in Figure 5.

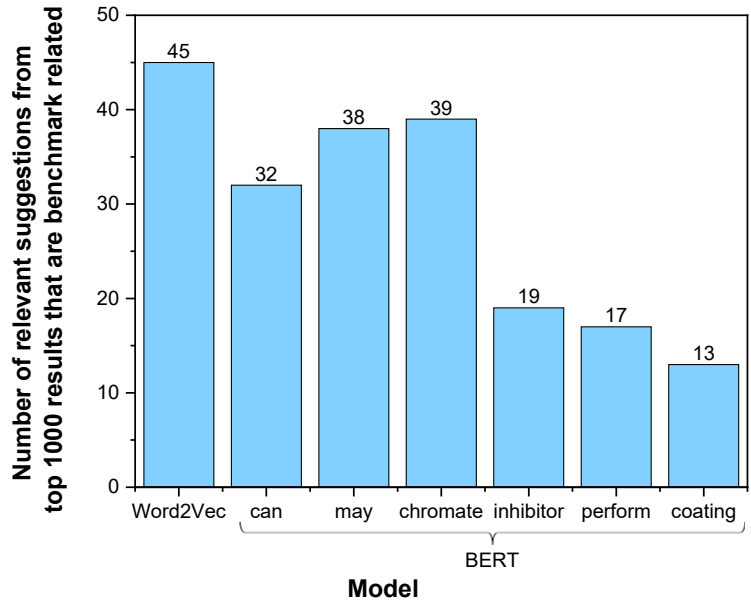

Figure 5. Number of benchmark alternatives materials in corrosion related results

Specifically, a total of 45 results (out of 54 relevant results overall) were considered as relevant benchmark alternatives from the Word2Vec model. The Word2Vec model therefore exhibited an 83.3% benchmark related rate, which was the highest rate - when compared to the six masked BERT models. It is also noted from Figure 5, that the first three masked sentences (from the BERT model) outperform the latter three masked sentences by a factor of nearly two.

To investigate these benchmark chromate alternative results in more detail, we focus on benchmark related results in each category predicted by Word2Vec and BERT model. Figure 6 reveals the count of benchmark related results in each of the twenty benchmark categories, isolating the performance of the Word2Vec model (in black) and the BERT model (in red). For this analysis, the BERT model is presented as a summation of the six masked models trialled, in order to examine overall performance.

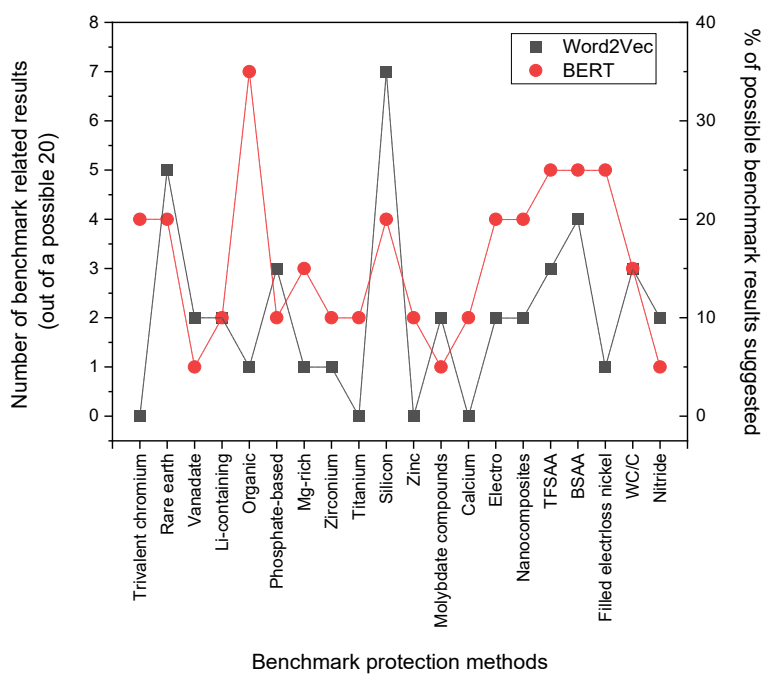

*Figure 6. Number of benchmark related predictions by Word2Vec and BERT model in 20 benchmark alternative category*

Inspection of Figure 6 reveals that the Word2Vec model did not identify four categories: trivalent chromium, titanium conversion coatings, zinc-based coatings and calcium-based coatings; while the BERT model covered all 20 categories, with at least one prediction. One of the categories, 'silicon-based systems', showed the highest number of predictions by the Word2Vec model. The other category of 'organic systems', revealed the same number of predictions by the BERT model. To further probe the four categories which were only identified by the BERT model, we report the prediction materials in each category and their frequency of occurrence in the original dataset, listed in Table 3. The frequency with which most of these materials was mentioned was relatively high, ranging from 25 to nearly 450 instances. We found that these words never appeared explicitly in the same sentence with "chromate", but they connected to "chromate" through other ways, such as "hydroxide" occurs in the same paragraph with "Cr", "TCP" and "chromium".

*Table 3. Four benchmark alternatives categories that only identified by BERT model*

| # | Category | Benchmark alternatives | Frequency |
|---|---|---|---|
| 1 | Trivalent chromium | fluoride, hydroxide, HF, ceramic | 25/52/24/291 |
| 2 | Titanium conversion coatings | titanium, Ti | 300/253 |
| 3 | Zinc-based coatings | zinc, Zn | 298/443 |
| 4 | Calcium based systems | calcium, Ca | 87/83 |

Overall, whilst the Word2Vec model BERT model revealed the highest benchmark related rate, that rate is only one metric of performance – and is directly linked to the number of predicted results. One of the more holistic assessments of model performance, when inspecting Figure 6, suggests that the BERT model outperformed the Word2Vec model for detecting all of the 20 benchmark chromate alternatives – including in the variety of approaches therein. Whilst not necessarily probed further in the present work, the ability of the BERT model to predict novel results, was also described. Conforming to the prediction of benchmark results – whilst meaningful in the initial exploration and validity of NLP approaches – is strong confirmation that expert human level interpretation is capable (in an automated process). From an aspirational perspective, the NLP approaches should extend beyond human level benchmarking and identify novel (as in, not readily having been interpreted by humans) results and correlations.

## General discussion

When comparing the predictions of six masked BERT models, it was noted that the input masked sentence structure significantly affects the performance of outputs. Since our goal herein was to search for possible chromate replacement, output noise data such as common words, verbs or adverbs should be avoided to the maximum extent. As a consequence, and based on initial work herein, it is posited that when the input masked word is a noun, this possibly correlates with better prediction performance (when comparing with expert human level benchmarks). As a way to analyse how the top 1000 chromate alternative predictions have identified corrosion-protection relevant results, we identified the corrosion relevant results from the Word2Vec and BERT models. It is acknowledged that the comparison between the two is not apples-and-apples; as the Word2Vec model runs as one/singular analysis; whereas the BERT model may be re-run numerous times (herein we used six masked models overall, and could have used more). This feature of the models tends to favour the BERT model, for 'user experience' and its ability to further probe for results. In regards to the latter point, here we revealed that the BERT model yielded more relevant results by almost three times, compared to the Word2Vec model. This finding suggests the BERT model, instead of focusing on the word "chromate", learns from context, builds instance-specific embedding for sentences, and therefore has a wider interpretation of corrosion protection relevant materials. However, when examining if the relevant results from the models tested are correlating with the benchmark related materials (Table 2), the Word2Vec model outperformed BERT, which is attributed to its word-sense for "chromate" and any known chromate replacement is expected to be 'highly related' to "chromate" and thus, have a similar word embedding. Whilst there is an ability to therefore capture what is already known as expert-human-level chromate alternatives from the literature using NLP (as demonstrated herein, and indeed by the Word2Vec model), this means that NLP can alleviate the need for a human to read and interpret hundreds (if into thousands) of papers. The task of becoming an expert can be replicated. However, the next step in suggesting novel alternatives for chromate, and nuance in context related interpretations, could be seen by the performance of the BERT model. The BERT model was able to predict certain benchmark alternatives that were not closely surrounded by "chromate" in the dataset. Additionally, BERT was capable of predicting low frequency and even zero frequency alternatives, - known as out-of-bag words. That is because BERT can generate vector representation in different sentences and there are infinite embeddings for each word type (and some of the out-of-bag words were indeed candidate alternatives). The study herein was an initial demonstration of NLP in the context of a corrosion challenge, an also, one of the first utilisations of the BERT model in a practical/applied engineering problem. Based on the work herein, there is significant scope for broadening the corpus for training, to include patents, websites, and other works – beyond the Scopus API, in the search for chromate alternatives.

## Conclusions

In this study, natural language processing (NLP) was utilized to automate the search of scientific literature for chromate replacements; specifically in the context of corrosion protection. It was revealed that the application of NLP was capable of serving in the role of searching for chromate replacements, without the need for a human to read any of the associated scientific literature. Herein, two unique NLP approaches were utilised, namely the Word2vec approach (previously explored in the field of materials by others) and the newer BERT approach, recently developed by Google. The latter approach was explored on the basis of its potential in handling out-of-vocabulary words, and its ability to operate by finding alternative words for a [mask] (i.e. the ability to 'answer questions asked' of the BERT model). Finding from the study herein can be summarised as:

- When comparing the NLP predictions from the work herein (which did not have a human in the loop) with three (3) benchmark studies/reviews from corrosion experts that have proposed a list of chromate replacements, it was determined that:
    - The Word2vec model predicted the most accurate chromate alternative results, by simply calculating the cosine distance.
    - The BERT model predicted the most extensive related results in the field, inclusive of even low frequency terms.

- - Both the BERT and Word2vec models could capture essentially all of the expert human determined chromate replacement technologies – albeit with no domain experience.
- NLP was able to readily capture scientific knowledge for a niche application, revealing the approaches employed herein – not developed for the application of chromate replacement – can serve as general approaches for broad applications
- Future work may explore the use of broader inputs, beyond those of the Scopus application programming interfaces, including webpages, and other collections. Specifically, broadening the inputs will possibly permit more novel chromate replacement predictions that are not in benchmark alternative list.

**Acknowledgements**

We gratefully acknowledge the funding from the Australian Research Council (ARC) through the Industrial Transformation Research Hubs Scheme under Project Number: IH200100005.

# Appendix 1

*Table A1: A ranked list of chromate alternatives from the utilisation of BERT model as tallied in Figure 4.*

| | can | | | may | | chromate | | inhibitor | | perform | | coating | |
|---|---|---|---|---|---|---|---|---|---|---|---|---|---|
| # | rank | prediction | rank | prediction | rank | prediction | rank | prediction | rank | prediction | rank | prediction |
| 1 | 3 | silica | 3 | nickel | 4 | hydroxide | 64 | coatings | 14 | primer | 163 | magnesium |
| 2 | 5 | zinc | 5 | zinc | 5 | carbon | 80 | stable | 18 | coatings | 224 | Al |
| 3 | 9 | titanium | 6 | cadmium | 9 | alumina | 167 | Co | 64 | inhibition | 343 | crystalline |
| 4 | 10 | nickel | 9 | titanium | 10 | HF | 220 | zinc | 91 | inhibitor | 355 | Co |
| 5 | 15 | cadmium | 10 | manganese | 11 | zinc | 257 | Zn | 141 | Co | 369 | Mo |
| 6 | 16 | phosphorus | 11 | magnesium | 13 | phosphate | 260 | organic | 158 | dimethyl | 385 | As |
| 7 | 17 | EDTA | 12 | phosphorus | 15 | chlorine | 274 | W | 184 | pigments | 420 | sol |
| 8 | 18 | magnesium | 14 | sulfur | 16 | sulfur | 309 | Mo | 207 | titanium | 443 | zinc |
| 9 | 20 | epoxy | 16 | silicon | 18 | manganese | 339 | cerium | 243 | In | 460 | amorphous |
| 10 | 21 | Co | 17 | fluoride | 20 | potassium | 340 | As | 295 | cobalt | 498 | Nb |
| 11 | 22 | fluoride | 18 | sodium | 22 | silica | 341 | Mg | 314 | phosphate | 504 | Se |
| 12 | 25 | polyethylene | 20 | cobalt | 24 | sulfate | 397 | sol | 333 | inhibiting | 520 | organic |
| 13 | 29 | hydroxide | 21 | selenium | 27 | alanine | 414 | Ni | 377 | epoxy | 539 | hybrid |
| 14 | 31 | primer | 24 | lithium | 29 | bicarbonate | 427 | alkaline | 385 | composite | 632 | alkaline |
| 15 | 34 | polymer | 26 | hydroxide | 30 | carbonate | 500 | V | 429 | CP | 647 | inhibited |
| 16 | 36 | clay | 27 | boron | 34 | nickel | 514 | Na | 454 | Mg | 673 | ceramic |
| 17 | 37 | HF | 28 | graphite | 35 | cobalt | 552 | Nb | 456 | alkaline | 697 | Mg |
| 18 | 47 | graphite | 29 | EDTA | 36 | phenol | 588 | S | 460 | V | 712 | Sb |
| 19 | 53 | CP | 30 | potassium | 37 | hydroxy | 594 | potassium | 461 | Sn | 717 | Ni |
| 20 | 55 | silicone | 34 | phosphate | 38 | NADH | 645 | sodium | 463 | silicon | 723 | RE |
| 21 | 56 | phosphate | 40 | silica | 39 | fluoride | 646 | pretreated | 492 | acetate | 735 | V |
| 22 | 57 | silicon | 41 | chlorine | 40 | phosphorus | 657 | inorganic | 503 | Ti | 770 | Ce |
| 23 | 60 | ceramic | 43 | arsenic | 41 | magnesium | 680 | nickel | 510 | sol | 805 | In |
| 24 | 64 | manganese | 46 | V | 44 | sodium | 720 | Au | 571 | ceramic | 810 | Sn |
| 25 | 66 | boron | 53 | HF | 46 | arginine | 754 | copolymer | 573 | hydroxide | 813 | Nd |
| 26 | 69 | glass | 58 | polyethylene | 53 | cadmium | 762 | clad | 584 | silica | 820 | sodium |
| 27 | 77 | cobalt | 66 | platinum | 60 | urea | 828 | Ni | 637 | benzotriazole | 829 | Pb |
| 28 | 80 | alanine | 67 | calcium | 62 | alkali | 829 | Ce | 649 | inorganic | 921 | glass |
| 29 | 81 | MEA | 82 | Zr | 63 | EDTA | 839 | antimony | 650 | amine | 923 | titanium |
| 30 | 85 | polysaccharides | 89 | mg | 64 | silicon | 842 | Rh | 655 | crystallization | 944 | silicon |
| 31 | 89 | nanoparticles | 90 | epoxy | 66 | steam | 845 | Pb | 657 | chlorine | | |
| 32 | 95 | WC | 91 | methylene | 69 | glycine | 890 | titanium | 706 | plating | | |
| 33 | 98 | CVD | 97 | nitrate | 70 | cysteine | 907 | ferric | 745 | hydroxyl | | |
| 34 | 99 | chlorine | 101 | Mo | 71 | graphite | 946 | Sn | 772 | K | | |
| 35 | 105 | Ta | 103 | Ti | 76 | boron | 976 | polymeric | 774 | oxides | | |
| 36 | 106 | NADH | 104 | zircon | 77 | ATP | 998 | alumina | 812 | zinc | | |
| 37 | 109 | sulfur | 105 | ferric | 83 | titanium | | | 845 | arginine | | |
| 38 | 112 | potassium | 106 | arginine | 85 | polysaccharides | | | 852 | Mn | | |
| 39 | 114 | platinum | 107 | alanine | 87 | glycerol | | | 905 | Ce | | |
| 40 | 117 | cysteine | 116 | Ni | 89 | ferric | | | 915 | citrate | | |
| 41 | 120 | citrate | 123 | alkaline | 90 | alkyl | | | 968 | nitrate | | |
| 42 | 122 | lithium | 137 | cysteine | 92 | clay | | | 980 | polymers | | |
| 43 | 146 | sol | 138 | phenol | 95 | hydroxyl | | | 986 | nickel | | |
| 44 | 153 | graphene | 142 | magnesium | 97 | Ti | | | | | | |
| 45 | 158 | urea | 149 | ethylene | 98 | Mg | | | | | | |
| 46 | 160 | arsenic | 150 | silicone | 107 | calcium | | | | | | |
| 47 | 170 | ferric | 158 | ceramic | 111 | amines | | | | | | |
| 48 | 187 | antibiotics | 160 | adenine | 114 | inorganic | | | | | | |
| 49 | 199 | ATP | 161 | bicarbonate | 116 | carbonyl | | | | | | |
| 50 | 208 | glycerol | 174 | WC | 118 | nitrate | | | | | | |
| 51 | 212 | arginine | 180 | carbonate | 124 | sulphate | | | | | | |
| 52 | 213 | methionine | 195 | NADH | 127 | lithium | | | | | | |
| 53 | 216 | sodium | 197 | quinone | 129 | casein | | | | | | |
| 54 | 217 | iodine | 202 | methionine | 130 | Mn | | | | | | |
| 55 | 237 | carbonate | 204 | ATP | 132 | pyridine | | | | | | |
| 56 | 246 | phenol | 224 | glycine | 139 | glutamine | | | | | | |
| 57 | 253 | crystallization | 226 | tris | 142 | graphene | | | | | | |
| 58 | 259 | hemoglobin | 257 | selenium | 143 | selenium | | | | | | |
| 59 | 260 | tetracycline | 275 | alkali | 145 | methionine | | | | | | |
| 60 | 270 | plating | 276 | graphene | 147 | Ca | | | | | | |
| 61 | 308 | selenium | 278 | crystallization | 150 | KOH | | | | | | |
| 62 | 309 | adenine | 297 | sulfate | 151 | adenine | | | | | | |
| 63 | 371 | bicarbonate | 301 | sulfide | 158 | citrate | | | | | | |
| 64 | 396 | Sb | 319 | pyridine | 161 | sulfide | | | | | | |
| 65 | 397 | Zr | 320 | phenolic | 162 | Y | | | | | | |
| 66 | 410 | organic | 336 | sulph | 165 | NADPH | | | | | | |
| 67 | 412 | acrylate | 341 | glycerol | 185 | formaldehyde | | | | | | |
| 68 | 414 | Ce | 354 | Ca | 188 | platinum | | | | | | |
| 69 | 428 | polyvinyl | 392 | glutamine | 200 | benzo | | | | | | |
| 70 | 439 | NADPH | 418 | acrylate | 210 | amine | | | | | | |
| 71 | 483 | Mn | 482 | urea | 218 | polyethylene | | | | | | |
| 72 | 541 | polypyrrole | 512 | polysaccharides | 222 | Na | | | | | | |
| 73 | 567 | composites | 519 | dimethyl | 227 | glutamate | | | | | | |
| 74 | 568 | amorphous | 668 | citrate | 250 | alkyl | | | | | | |
| 75 | 602 | Zr | 787 | amine | 255 | silicone | | | | | | |
| 76 | 655 | alkali | 794 | vitamin | 263 | arsenic | | | | | | |
| 77 | 772 | amine | 811 | NADPH | 308 | Zr | | | | | | |
| 78 | 922 | tris | 816 | amines | 341 | surfactants | | | | | | |
| 79 | 929 | pseudomonas | 900 | Phosphate | 390 | Co | | | | | | |
| 80 | | | 946 | Arsenate | 411 | epoxy | | | | | | |
| 81 | | | | | 420 | amine | | | | | | |
| 82 | | | | | 437 | alkaline | | | | | | |
| 83 | | | | | 688 | nitrite | | | | | | |
| 84 | | | | | 875 | nanoparticles | | | | | | |
| 85 | | | | | 880 | pseudomonas | | | | | | |

## Appendix 2

Table A2: A ranked list of chromate alternatives from the utilisation of the Word2Vec model as tallied in Figure 4.

| # | Rank | Word2Vec |
|---|---|---|
| 1 | 6 | silane |
| 2 | 10 | molybdate |
| 3 | 36 | sol-gel |
| 4 | 37 | PANI |
| 5 | 42 | cerium |
| 6 | 47 | Mg-rich |
| 7 | 49 | sol-gel coatings |
| 8 | 51 | polyaniline |
| 9 | 62 | lithium |
| 10 | 67 | electroplating |
| 11 | 79 | anodising |
| 12 | 80 | zinc phosphate |
| 13 | 90 | anodizing |
| 14 | 109 | epoxy |
| 15 | 128 | sol-gel |
| 16 | 130 | Cerium nitrate |
| 17 | 223 | PVD |
| 18 | 228 | $ZrO_2$ |
| 19 | 229 | hard chromium plating |
| 20 | 239 | sulfuric |
| 21 | 268 | tungsten carbide |
| 22 | 302 | silica |
| 23 | 305 | graphene |
| 24 | 309 | diamond-like |
| 25 | 310 | sol |
| 26 | 332 | phosphates |
| 27 | 352 | Ni-P |
| 28 | 370 | organosilane |
| 29 | 410 | WC–Co |
| 30 | 454 | $MoNa_2O_4$ |
| 31 | 462 | $BiO_4V$ |
| 32 | 491 | Ce-Nd |
| 33 | 495 | Co-P |
| 34 | 512 | CNT |
| 35 | 516 | vanadate |
| 36 | 545 | phytic acid |
| 37 | 552 | neodymium |
| 38 | 584 | silicate |
| 39 | 585 | $CeN_3O_9$ |
| 40 | 596 | Ni–P |
| 41 | 628 | lanthanum |
| 42 | 657 | clays |
| 43 | 660 | sulphuric |
| 44 | 666 | magnesium |
| 45 | 693 | $Cr_xN$ |
| 46 | 732 | Ce |
| 47 | 734 | SAA |
| 48 | 765 | WC-(WCr)$_2$C-Ni |
| 49 | 787 | polypyrrole |
| 50 | 800 | silicon |
| 51 | 866 | phosphate |
| 52 | 892 | LDHs |
| 53 | 909 | $CrN_x$ |
| 54 | 923 | Nd |